\documentclass[journal]{IEEEtran}
\ifCLASSINFOpdf
   \usepackage[pdftex]{graphicx}
\else
   \usepackage[dvips]{graphicx}
\fi
\begin{document}
\title{Low-Rank Discriminative Least Squares Regression for Image Classification}

\author{Zhe~Chen,~\IEEEmembership{}
        Xiao-Jun~Wu$^*$,~\IEEEmembership{}      
        and~Josef~Kittler,~\IEEEmembership{Life~Member,~IEEE}% <-this % stops a space
\IEEEcompsocitemizethanks{\IEEEcompsocthanksitem

$^*$Corresponding author. 

E-mail: wu\_xiaojun@jiangnan.edu.cn  

Zhe Chen and Xiao-Jun Wu are with the School of Internet of Things, Jiangnan University, Wuxi 214122, China.

Josef Kittler is with the Centre for Vision, Speech and Signal Processing (CVSSP), University of Surrey, Guildford GU2 7XH, U.K.
 }% <-this % stops an unwanted space
\thanks{}}


\markboth{Journal of \LaTeX\ Class Files}%
{}



\maketitle

\begin{abstract}
Latest least squares regression (LSR) methods aim to learn slack regression targets to replace strict zero-one labels. However, the difference between intra-class targets can also be highlighted when enhancing the distance between different classes, and roughly persuing relaxed targets may lead to the problem of overfitting. To solve above problems, we propose a low-rank discriminative least squares regression model (LRDLSR) for multi-class image classification. Specifically, LRDLSR class-wisely imposes low-rank constraint on the intra-class regression targets to encourage its compactness and similarity. Moreover, LRDLSR introduces an additional regularization term on the learned targets to avoid the problem of overfitting. We show that these two improvements help to learn a more discriminative projection for regression, thus achieving better classification performance. The experimental results over a range of image databases demonstrate the effectiveness of the proposed LRDLSR method.
\end{abstract}

\begin{IEEEkeywords}
least squares regression, low-rank regression targets, overfitting, image classification 
\end{IEEEkeywords}

% \fi
\IEEEpeerreviewmaketitle

\section{Introduction}
\IEEEPARstart{L}{east} squares regression (LSR) is a very popular method in the field of multicategory image classification. LSR aims at learning a projection to transform the original data into the corresponding zero-one labels with a minimum loss. Over the past decades, many LSR based variants have been developed, such as locally weighted LSR \cite{LWLSR}, local LSR \cite{LLSR},  LASSO regression \cite{LASSO}, kernel ridge LSR \cite{KRLSR}, kernel LSR \cite{KLSR}, weighted LSR \cite{WLSR}, a least-squares support vector machine (LS-SVM) \cite{LSSVM} and partial LSR \cite{PLSR}. Besides linear regression, sparse representation, collaborative representation, and probabilistic collaborative representation based classification methods (LRC, SRC, CRC and ProCRC) \cite{LRC}\cite{SRC}\cite{CRC}\cite{ProCRC} also take advantage of the LSR framework to find representation coefficients.

However, there are still many issues associated with the above LSR based methods. First, taking the zero-one label matrix as the regression targets is too strict. It is not ideal for classification, as calculating the least squares loss between the extracted features and binary targets cannot reflect the classification performance of a regression model, especially in the multi-class conditions. For instance, the Euclidean distance of any two of the inter-class regression targets is constant, i.e., $ \sqrt{2}$, and  for each sample, the difference between the targets of the true and the false class identically equals to 1.  These characteristics are contrary to the expectation that the transformed inter-class features should be as far as possible from each other. To address this problem, some representative algorithms, i.e., discriminative LSR (DLSR) \cite{DLSR}, retargeted LSR (ReLSR) \cite{ReLSR},  and groupwise ReLSR \cite{GReLSR}, were proposed to learn relaxed regression targets instead of the original binary targets. Concretely, DLSR utilizes the $\varepsilon$-dragging technique to encourage the inter-class regression targets moving in the opposite directions, thus enlarging the distances between different classes. Different from DLSR, ReLSR learns the regression targets from the original data rather than directly adopting the zero-one labels of the samples, in which the margins between classes are forced to be greater than 1. Lately, Wang et al. \cite{ReLSR} proved that DLSR is a special model of ReLSR, with the translation values set to zero, and proposed a new formulation for ReLSR. With the new formulation, GReLSR introduces a groupwise constraint to guarantee that intra-class samples have similar translation values.

Besides, the traditional LSR based methods also do not take into account the data correlation during the projection learning procedure, which may result in the loss of some useful structural information and cause overfitting. To explore the underlying relationships, Fang \emph{et al.} \cite{RLRLR} constructed a class-compactness-graph to ensure that the projected intra-class features are compact so that the overfitting problem can be mitigated to some degree. Wen \emph{et al.} \cite{ICS_DLSR} proposed a novel framework called inter-class sparsity based DLSR (ICS\_DLSR), which introduces an inter-class sparsity constraint on the DLSR model to make the projected features of each class retain sparse structure. In fact, both, the RLRLR and ICS\_DLSR algorithms, are based on the model of DLSR, that is they adopt the $\varepsilon$-dragging technique. In addition to learning slack regression targets, RLSL \cite{RLSL} proposed to jointly learn the latent feature subspace and classification model so that the data representation extracted is more dicriminative and compact for classification. The learned latent subspace can be regarded as a transition between the original samples and binary labels. 

The various measures adopted by the algorithms mentioned above improve the classification performance. However the $\varepsilon$-dragging technique or the margin constraint used to relax the label matrix also amplify the difference among the intra-class regression targets, which may deteriorate the classification performance. In this paper, a novel relaxed targets based regression model named low-rank discriminative least squares regression (LRDLSR) is proposed to learn a more discriminative projection. Based on the model of DLSR, LRDLSR class-wise imposes a low-rank constraint on the relaxed regression targets to ensure the intra-class targets are compact and similar. In this way, the $\varepsilon$-dragging technique will be exploited to a better effect so that both the intra-class similarity and the inter-class seperability of regression targets can be guaranteed. Moreover, LRDLSR minimizes the energy of the resulting dynamic regression targets to avoid the problem of overfitting.

The rest of this paper is organized as follows. First, the related works are briefly introduced in Section II. The proposed LRDLSR model and the corresponding optimization procedure are described in Section III. The properties of the algorithm are analysed in Section IV. The experimental results are presented in Section V and Section VI concludes this paper.

\section{Related works}
In this section, we briefly review the related works. Let $X=[x_1, x_2,..., x_n]\in R^{d\times n}$ denote the $n$ training samples from $c$ classes $(c\geq 2)$, where $d$ is the dimensionality of the samples. $X_i\in R^{d\times n_i}$ denotes the subset of the samples belonging to the $i$th class. $H=[h_1, h_2,..., h_n]\in R^{c\times n}$ denote the binary label matrix of $X$, where column $h_i$ of $H$, i.e., $h_i=[0,...,0,1,0,...,0]^T\in R^c$, corresponds to the training sample $x_i$. If sample $x_i$ belongs to the $p$th class,  then the $p$th element of $h_i$ is 1 and all the others are 0.

\subsection{Original LSR}
The main idea of LSR is to learn a projection matrix that maps the original training samples into the binary label space. The objective function of LSR can be formulated as
$$\min_Q \|QX-H\|_F^2+\lambda\|Q\|_F^2 \eqno{(1)}$$
where $\|\bullet\|_F^2$ is the matrix Frobenius norm $(\|A\|_F^2=tr(A^TA)=tr(AA^T))$ and $\lambda$ is a positive regularization parameter. $Q$ is the projection matrix. The first term in problem (1) is a least squares loss function, while the second term is used to avoid the problem of overfitting. Obviously, (1) has a closed-form solution as 
$$\hat{Q} = HX^T(XX^T + \lambda I)^{-1} \eqno{(2)} $$

Given a new sample $y$, LSR calculates its label as 
$k = arg\max_j(Qy)_j $
where $(Qy)_j$ is the $j$th value of $Qy$.

\subsection{DLSR and ReLSR}
As previously said, making the regression features to pursue strict zero-one outputs is inappropriate for classification tasks. Unlike original LSR, DLSR \cite{DLSR} and ReLSR \cite{ReLSR} aim at learning relaxed regression targets rather than using the binary labels $H$ as their targets. The main idea of DLSR is to
enlarge the distance between the true and the false classes by using an $\varepsilon$-dragging technique. Its regression model can be formulated as
$$\min_{Q,M} \|QX-(H+B\odot M)\|_F^2+\lambda\|Q\|_F^2, \ s.t. \ M\geq 0 \eqno{(3)}$$
where $\odot$ denotes the Hadamard-product operator. $M\in R^{c\times n}$ is a non-negative $\varepsilon$-dragging label relaxation matrix. $B\in R^{c\times n}$ is a constant matrix which is defined as
\setcounter{equation}{3}
\begin{eqnarray}
B_{ij} = \left\{
\begin{array}{lcl}
+1, & if & H_{ij}=1 \\
-1, & if & H_{ij}=0 \\  
\end{array}
\right. 
\end{eqnarray}

Compared to the original LSR, it can be seen that the regression targets are extended to be $H'=H+B\odot M$ in DLSR. To help the understanding, we use four samples to explain why the new relaxed target matrix $H'$ is more discriminative than $H$. Let $x_1$, $x_2$, $x_3$, $x_4$ be four training samples in which the first two samples are from the first class and the latter two are from the second class. Thus their binary label matrix is defined as 
\setcounter{equation}{4}
\begin{equation}       
H=\left[                
  \begin{array}{cccc}   
    1 & 1 & 0 & 0\\  
    0 & 0 & 1 & 1  
  \end{array}
\right]    \in R^{2\times4}             
\end{equation}
It is obvious that the distance between any two inter-class targets is $\sqrt{2}$ $(\sqrt{(1-0)^2+(0-1)^2}=\sqrt{2})$. Such a fixed distance cannot reflect the classification ability of regression model well. But  if we use $H'$ to replace $H$, then we have 
\setcounter{equation}{5}
\begin{equation}       
H'=\left[                
  \begin{array}{cccc}   
    1+m_{11} & 1+m_{12} & -m_{13} & -m_{14}\\  
    -m_{21} & -m_{22} & 1+m_{23} & 1+m_{24}  
  \end{array}
\right]       
\end{equation}
In doing so, the distance between the first and the fourth target is $\sqrt{(1+m_{11}+m_{14})^2+(-m_{21}-1-m_{24})^2}\geq\sqrt{2}$ rather than a constant. The margin between the two classes is also enlarged by changing the regression outputs in the opposite directions. For example, the class margin of the first regression target is $1+m_{11}+m_{21}>1$. These meet the expectation that inter-class samples
should be as far as possible from each other after being projected.

Likewise, ReLSR directly learns relaxed regression targets from the original data to ensure that samples are correctly classified with large margins. The ReLSR model is defined as
\setcounter{equation}{6} 
\begin{eqnarray}
\min_{Q,T,b}\|T-QX-be_n\|_F^2+\lambda\|Q\|_F^2  \nonumber \\
 s.t. T_{r_j,j}-\max_{i\neq r_j}T_{i,j}\geq1 , i=1,2,....,n
\end{eqnarray}
where $r_j$ indicates the true label of sample $x_j$. $T$ is optimized from $X$ with a large margin constraint which enhances the class separability. Hence ReLSR performs better flexibility than DLSR.

\section{From DLSR and ReLSR to LRDLSR}
\subsection{Problem Formulation and New Regression Model}
Although DLSR and ReLSR can learn soft targets and maintain the closed-form solution for the projection, an undue focus on large margins will also result in overfitting. As indicated before, exploiting the data correlations is helpful in learning a discriminative data representation. From the classification point of view, both, the intra-class similarity and the inter-class incoherence of regression targets should be promoted. However, DLSR and ReLSR ignore the former, because their relaxation values are dynamic. Hence the $\varepsilon$-dragging technique in DLSR and the margin constrain method in ReLSR will also promote the intra-class regression targets to be discrete. If the intra-class similarity of learned targets is weakened, the discriminative power will be compromised. Therefore, based on the model of DLSR, we propose a low-rank discriminative least squares regression model (LRDLSR) as follows 
\setcounter{equation}{7}  
\begin{eqnarray}
\min_{Q,T,M}\frac{1}{2}\|QX-T\|_F^2+\frac{\alpha}{2}\|T-(H+B\odot M)\|_F^2+\nonumber \\ 
\beta\sum_{i=1}^c\|T_i\|_*+ \frac{\gamma}{2}\|T\|_F^2+\frac{\lambda}{2}\|Q\|_F^2,\ s.t. M\geq 0
\end{eqnarray}
where $\alpha$, $\beta$, $\gamma$ and $\lambda$ are the regularization parameters and $\|\bullet\|_*$ denotes the nuclear norm (the sum of singular values). $Q$ and $T$ denote the projection matrix and the slack target matrix, respectively. The second term $\|T-(H+B\odot M)\|_F^2$ is used to learn relaxed regression targets with large inter-class margins, the third term $\sum_{i=1}^c\|T_i\|_*$ is used to learn similar intra-class regression targets, and the fourth term $\|T\|_F^2$ is used to avoid the overfitting problem of $T$. 

With our formulation, we note that the major difference between our LRDLSR and DLSR is that in LRDLSR we encourage the relaxed regression targets of each class to be low-rank so that the compactness and similarity of the regression targets from each class can be enhanced. Combined with the $\varepsilon$-dragging technique, both the intra-class similarity and inter-class separability of regression targets will be preserved, thus producing a discriminative projection. In fact, the proposed class-wise low-rank constraint term can also be extended to the ReLSR and GReLSR models, or other relaxed target learning based LSR models. In addition to the above difference, we also add a simple $F$-norm constraint on $T$, i.e., $\|T\|_F^2$, to restrict the energy of the targets $T$. This is because there are no any restrictions on the variation magnitude of the dynamically updated regression targets in DLSR. In this way, the slack matrix, i.e., $M$, may be very fluctuant and discrete because of aggressively exploiting the largest class margins, thus leading to the problem of overfitting.

\subsection{Optimization of LRDLSR}
To directly solve the optimization problem in (8) is impossible because three variables $Q$, $T$ and $M$ are correlated. Therefore, an iterative update rule is devised to solve it so as to guarantee that it has a closed-form solution in each iteration. In this paper, the alternating direction multipliers method (ADMM) \cite{ADMM} is exploited to optimize LRDLSR. In order to make (8) separable,  we first introduce an auxiliary variable $P$ as follows
\setcounter{equation}{8}  
\begin{eqnarray}
\min_{T,P,Q,M}\frac{1}{2}\|QX-T\|_F^2+\frac{\alpha}{2}\|T-(H+B\odot M)\|_F^2+\nonumber \\ 
\beta\sum_{i=1}^c\|P_i\|_*+ \frac{\gamma}{2}\|T\|_F^2+\frac{\lambda}{2}\|Q\|_F^2,\ s.t.\ T=P, M\geq 0
\end{eqnarray}
Then we obtain the augmented Lagrangian function of (9) 
\setcounter{equation}{9}  
\begin{eqnarray}
L(T,P,Q,M,Y)=\frac{1}{2}\|QX-T\|_F^2+\nonumber \\ 
\frac{\alpha}{2}\|T-(H+B\odot M)\|_F^2+\beta\sum_{i=1}^c\|P_i\|_*+ \frac{\gamma}{2}\|T\|_F^2+\nonumber \\\frac{\lambda}{2}\|Q\|_F^2+\frac{\mu}{2}\|T-P+\frac{Y}{\mu}\|_F^2
\end{eqnarray}
where $Y$ is the Lagrangian multiplier, $\mu>0$ is the penalty parameter. Next we update variables one by one.

\noindent\textbf{Update $T$:} By fixing variables $P$, $Q$, $M$, $T$ can be obtained by minimizing the following problem
\setcounter{equation}{10}  
\begin{eqnarray}
L(T)=\frac{1}{2}\|QX-T\|_F^2+\frac{\alpha}{2}\|T-(H+B\odot M)\|_F^2+\nonumber \\
\frac{\gamma}{2}\|T\|_F^2+\frac{\mu}{2}\|T-P+\frac{Y}{\mu}\|_F^2 
\end{eqnarray}

Obviously, $T$ has a closed-form solution as
$$T=(1+\alpha+\gamma+\mu)^{-1}[QX+\alpha(H+B\odot M)+\mu P - Y] \eqno{(12)}$$

\noindent\textbf{Update $P$:} Given $T$, $Q$ and $M$, $P$ can be class-wisely updated by
$$L(P_i)=\beta\sum_{i=1}^c\|P_i\|_*+ \frac{\mu}{2}\|T_i-P_i+\frac{Y_i}{\mu}\|_F^2\eqno{(13)}$$

We can use the singular value thresholding algorithm \cite{SVD} to class-wisely optimize (13). The optimal solution of $P_i$ is 
$$P_i=I_{\frac{\beta}{\mu}}(T_i+\frac{Y_i}{\mu}) \eqno{(14)}$$
where $I_\zeta(\Theta)$ is the singular value shrinkage operator.

(\uppercase\expandafter{\romannumeral1}) Given a matrix $\Theta \in R^{a\times b}$, its singular value decomposition can be formulated as 
$$\Theta=U_{a\times r}\Sigma V_{b\times r}^T, \quad \Sigma=diag(\sigma_1,...,\sigma_r). \eqno{(15)}$$
where $r$ is the rank of $\Theta$, $U$ and $V$ are column-orthogonal matrices.

(\uppercase\expandafter{\romannumeral2}) Given a threshold $\zeta$,
$$I_\zeta(\Theta)=U_{a\times r}diag(\{max(0,\sigma_j-\zeta)\}_{1\leq j\leq r})V_{b\times r}^T. \eqno{(16)}$$

\noindent\textbf{Update $Q$:} Analogously, $Q$ can be solved by minimizing
$$L(Q)=\frac{1}{2}\|QX-T\|_F^2+\frac{\lambda}{2}\|Q\|_F^2 \eqno{(17)}$$

We set the derivative of $L(Q)$ with respect to $Q$ to zero, and obtain the following closed-form solution
$$Q=TX^T(XX^T+\lambda I)^{-1}\eqno{(18)}$$
Let $R=X^T(XX^T+\lambda I)^{-1}$, we find that $R$ is independent of $T$, thus $R$ can be pre-calculated before starting the iteration.

\noindent\textbf{Update $M$:} After optimizing $T$, $P$ and $Q$, we can update the non-negative relaxation matrix $M$ by
$$\min_M\|T-(H+B\odot M)\|_F^2,\ s.t.\ M\geq 0 \eqno{(19)}$$

Let $R=T-H$, according to \cite{DLSR}, the optimal solution of $M$ can be calculated by
$$M=max(B\odot R,\ 0) \eqno{(20)}$$

The optimization procedure of LRDLSR is overviewed in Algorithm 1.  
%算法1
\begin{table}[!ht]
\rule[0.1cm]{8.8cm}{1.5pt}
\leftline {\textbf {Algorithm 1.} Optimizing LRDLSR by ADMM}\\
\rule[0.1cm]{8.8cm}{1.5pt}
\textbf{Input:} Normalized training samples $X$ and its label matrix $H$; Parameters $\alpha, \beta, \gamma, \lambda$.

\textbf{Initialization:} $T=P=H$, $Q=\textbf0$, $M=\textbf1^{c\times n}$, $Y=\textbf0^{c\times n}$, $\mu_{max}=10^{8}$, $tol=10^{-6}$, $\mu=10^{-5}$, $\rho=1.1$.\\
\textbf{While} not converged do:
 \begin{enumerate}
\item Update $T$ by using Eq. (12).
\item Update $P$ by using Eq. (13).
\item Update $Q$ by using Eq. (18).
\item Update $M$ by using Eq. (20).
\item Update Lagrange multipliers $Y$ as
$$Y=Y+\mu(T-P). \eqno{(17)}$$
\item Update penalty parameter $\mu$ as
$$\mu=min(\mu_{max}, \rho \mu). \eqno{(18)}$$
\item Check convergence: 
$$if \ \|T-P\|_{\infty}\leq tol. \eqno{(19)}$$
\end{enumerate}
\textbf{End While}\\
\textbf{Output:} $Q, T$ and $M$.\\
\rule[0.1cm]{8.8cm}{1.5pt}
\end{table}

\subsection{Classification}
Once (8) is solved, we can obtain the optimal projection matrix $Q$. Then, we use $Q$ to obtain the projection features of the training samples, i.e., $QX$. Suppose $y\in R^d$ is a test sample, then its projection feature is $Qy$. For convenience, we use the NN classifier to implement classification in our paper.

\section{Analysis of the proposed method}

\subsection{Computational Complexity}
In this section, we analyze the computational complexity of Algorithm 1. The main time-consuming steps of Algorithm 1 are

(1) Singular value decomposition in Eq. (13).

(2) Matrix inverse in (18).

Since the remaining steps only consist of simple matrix addition, substraction and multiplication operations, and element-wise multiplication operation, similar to \cite{ICS_DLSR}\cite{SRNMR}, we also ignore the time complexity of these operations.  
The complexity of singular value decomposition in Eq. (13) is $O(min(cn_i^2,n_ic^2))$. The complexity of pre-computing $(XX^T+\lambda I)^{-1}$ in Eq. (14) is $O(d^3)$. Thus the final time complexity for Algorithm 1 is about $O(d^3 + \tau \sum_{i=1}^c min[cn_i^2, c^2n_i)]$, where $\tau$ is the number of iterations.

\subsection{Convergence validation}
In this section, we experimentally validate the convergence property of the ADMM optimization algorithm. Fig. 1 gives an empirical evidence that Algorithm 1 converges very well. The value of objective function monotonically decreases with the increasing number of iterations in four different databases. This indicates the effectiveness of the optimization method. However, it is still arduous to theoretically demonstrate that our optimization algorithm has strong convergence because problem (8) has four different blocks and the overall model of LRDLSR is non-convex. 
%图1 convergence analysis of LRDLSR

\begin{figure}[!h]
\centering
\includegraphics[scale=0.32]{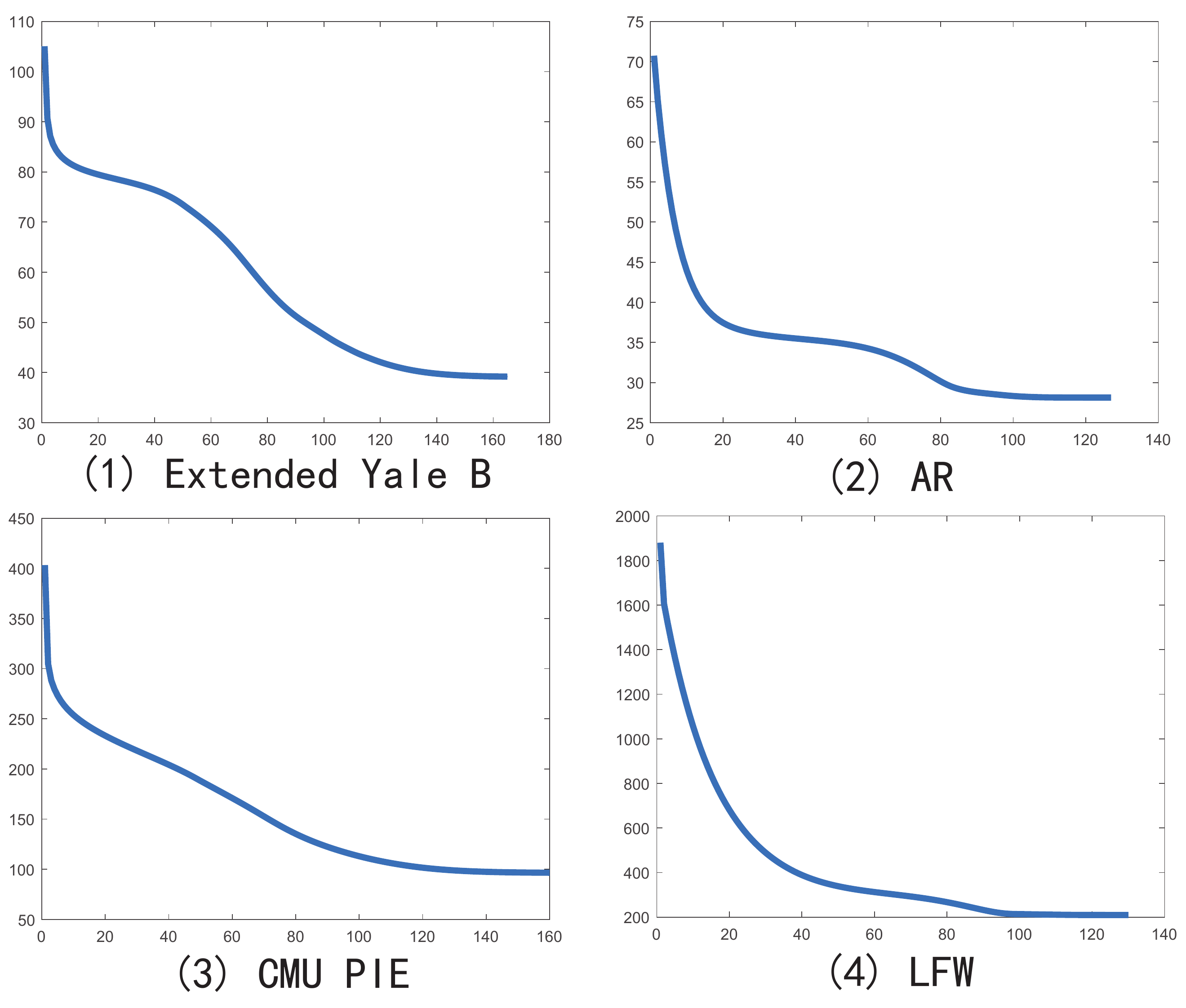}
\caption{Convergence curves and objective function versus iterations of LRDLSR on four face datasets.}
\end{figure}

\section{Experiments}
In order to verify the effectiveness of the proposed LRDLSR model, we compare it with five state-of-the-art LSR model based classification methods, including DLSR \cite{DLSR}, ReLSR \cite{ReLSR}, GReLSR \cite{GReLSR}, RLRLR\cite{RLRLR} and  RLSL\cite{RLSL}, and three representation based classification methods, including LRC \cite{LRC}, CRC \cite{CRC}, ProCRC \cite{ProCRC}, on five real image datasets. For LRDLSR, DLSR, ReLSR, GReLSR, RLRLR and RLSL, we use the NN classifier. For RLSL, the parameter $d$ is set to $2c$, where $c$ is the number of classes. When we test the performance of CRC, LRC and ProCRC,  all the training samples are used as the dictionary. To make fair comparisons, we directly utilize the released codes of the methods being compared to conduct experiments and seek the best parameters for them as much as possible. All the experiments are repeated ten times with random splits of training and test samples. The average results and the standard deviations (mean$\pm$std) are reported. The image datasets used in our experiments can be divided into two types:

(1) Face: the AR \cite{AR}, the CMU PIE \cite{CMU_PIE}, the Extended Yale B \cite{EYB} and the Labeled Faces in the Wild (LFW) \cite{LFW} datasets; 

(2) Object: the COIL-20 \cite{COIL} dataset.

%表1 
\begin{table*}[!t]
\renewcommand{\arraystretch}{1.3}
\caption{Recognition accuracy (mean$\pm$ std\%) of different methods on the COIL-20 object database.}
\label{table_example}
\centering
\begin{tabular}{|c|c|c|c|c|c|c|c|c|c|}
\hline
Train No. & LRC & CRC & ProCRC & DLSR & ReLSR & GReLSR & RLRLR & RLSL & LRDLSR(ours)\\
\hline
10 & 92.30$\pm$1.15 & 89.09$\pm$1.48 & 90.61$\pm$0.95 & 93.27$\pm$1.43 & 93.65$\pm$1.94 & 90.98$\pm$1.62 & 92.61$\pm$1.04 & 94.80$\pm$1.16 & \textbf{95.12$\pm$1.22}\\
\hline
15 & 94.89$\pm$1.33 & 92.58$\pm$1.27 & 94.53$\pm$0.85 & 96.25$\pm$0.75 & 96.75$\pm$0.72 & 93.60$\pm$0.83 & 94.86$\pm$0.85 &96.09$\pm$0.90 & \textbf{97.78$\pm$0.86}\\
\hline
20 & 97.49$\pm$0.51 & 94.15$\pm$1.15 & 96.17$\pm$0.82 & 97.52$\pm$0.67 & 98.17$\pm$0.67 & 95.65$\pm$0.82 & 96.27$\pm$0.33 & 97.45$\pm$0.29 & \textbf{98.51$\pm$0.85}\\
\hline
25 & 98.32$\pm$0.60 & 94.99$\pm$1.24 & 97.53$\pm$0.68 & 98.67$\pm$0.53 & 98.90$\pm$0.85 & 96.30$\pm$0.84 & 96.76$\pm$1.06 & 97.66$\pm$0.91 & \textbf{99.24$\pm$0.59}\\
\hline
\end{tabular}
\end{table*}

%表2 
\begin{table*}[]
\renewcommand{\arraystretch}{1.3}
\caption{Recognition accuracy (mean$\pm$ std\%) of different methods on the Extended Yale B face database.}
\label{table_example}
\centering
\begin{tabular}{|c|c|c|c|c|c|c|c|c|c|}
\hline
Train No. & LRC & CRC & ProCRC & DLSR & ReLSR & GReLSR & RLRLR & RLSL & LRDLSR(ours)\\
\hline
10 & 82.18$\pm$0.92 &   \textbf{91.85$\pm$0.61} & 91.74$\pm$0.86 & 87.95$\pm$1.10 & 89.68$\pm$0.94 & 88.46$\pm$1.00 & 90.21$\pm$0.84 & 89.02$\pm$0.88 & 91.18$\pm$0.65\\
\hline
15 & 89.43$\pm$0.58 &   94.76$\pm$0.66 & \textbf{95.41$\pm$0.76} & 93.37$\pm$0.99 & 93.98$\pm$0.52 & 93.13$\pm$0.82 & 94.80$\pm$0.64 & 93.29$\pm$0.73 & 95.07$\pm$0.66\\
\hline
20 & 92.00$\pm$0.77 &   96.39$\pm$0.56 & 96.74$\pm$0.26 & 95.73$\pm$0.68 & 96.14$\pm$0.54 & 95.25$\pm$0.50 & 96.37$\pm$0.71 & 95.18$\pm$0.62 & \textbf{96.84$\pm$0.36}\\
\hline
25 & 93.73$\pm$0.79 &  97.69$\pm$0.40 & 97.58$\pm$0.37 & 97.34$\pm$0.55 & 97.75$\pm$0.64 & 97.06$\pm$0.37 & 97.34$\pm$0.50 & 96.69$\pm$0.63 & \textbf{98.16$\pm$0.46}\\
\hline
\end{tabular}
\end{table*}

%表3 
\begin{table*}[]
\renewcommand{\arraystretch}{1.3}
\caption{Recognition accuracy (mean$\pm$ std\%) of different methods on the AR face database.}
\label{table_example}
\centering
\begin{tabular}{|c|c|c|c|c|c|c|c|c|c|}
\hline
Train No. & LRC & CRC & ProCRC & DLSR & ReLSR & GReLSR & RLRLR & RLSL & LRDLSR(ours)\\
\hline
3 & 28.73$\pm$0.99 & 71.42$\pm$0.59 & 76.16$\pm$1.12 & 73.58$\pm$1.63 & 73.53$\pm$1.47 & 74.77$\pm$1.45 &76.39$\pm$1.56 &75.70$\pm$1.01 & \textbf{78.80$\pm$0.76}\\
\hline
4 & 37.21$\pm$1.13 & 78.50$\pm$0.67 & 83.58$\pm$0.82 & 80.47$\pm$1.36 & 81.46$\pm$0.79 & 82.54$\pm$1.24 & 83.55$\pm$1.35 & 83.02$\pm$0.79& \textbf{86.20$\pm$0.45}\\
\hline
5 & 44.69$\pm$1.22 & 83.54$\pm$0.67 & 87.33$\pm$0.74 & 85.33$\pm$0.93 & 86.43$\pm$0.94 & 87.35$\pm$1.21 & 86.68$\pm$0.54 & 86.37$\pm$0.40 & \textbf{90.16$\pm$0.75}\\
\hline
6 & 52.95$\pm$1.54 & 86.79$\pm$0.71 & 90.32$\pm$0.66 & 88.18$\pm$0.78 & 88.98$\pm$0.99 & 89.96$\pm$0.73 &89.41$\pm$0.89 &88.80$\pm$0.48 & \textbf{92.23$\pm$0.80}\\
\hline
\end{tabular}
\end{table*}

 %表4 
\begin{table*}[]
\renewcommand{\arraystretch}{1.3}
\caption{Recognition accuracy (mean$\pm$ std\%) of different methods on the CMU PIE face database.}
\label{table_example}
\centering
\begin{tabular}{|c|c|c|c|c|c|c|c|c|c|}
\hline
Train No. & LRC & CRC & ProCRC & DLSR & ReLSR & GReLSR & RLRLR & RLSL & LRDLSR(ours)\\
\hline
10 & 75.67$\pm$1.01  & 86.39$\pm$0.60 & 89.00$\pm$0.37 & 87.54$\pm$0.79 & 88.18$\pm$0.79 & 86.88$\pm$0.72 &91.15$\pm$0.58 & 87.70$\pm$0.63 & \textbf{91.57$\pm$0.48}\\
\hline
15 & 85.26$\pm$0.63 & 91.14$\pm$0.43 & 92.18$\pm$0.25 & 92.22$\pm$0.54 & 92.29$\pm$0.42 & 91.21$\pm$0.51 &93.52$\pm$0.32 & 91.38$\pm$0.43 & \textbf{94.45$\pm$0.51}\\
\hline
20 & 89.84$\pm$0.48 & 93.08$\pm$0.35 & 93.94$\pm$0.18 & 94.12$\pm$0.27 & 94.23$\pm$0.21 & 93.39$\pm$0.27 & 94.78$\pm$0.30 & 93.03$\pm$0.38 & \textbf{95.83$\pm$0.35}\\
\hline
25 & 92.55$\pm$0.39 & 94.12$\pm$0.30 & 94.58$\pm$0.21 & 95.25$\pm$0.20 & 95.53$\pm$0.16 & 94.32$\pm$0.31 & 95.40$\pm$0.18 & 94.04$\pm$0.27 & \textbf{96.59$\pm$0.21}\\
\hline
\end{tabular}
\end{table*}

%表5 
\begin{table*}[]
\renewcommand{\arraystretch}{1.3}
\caption{Recognition accuracy (mean$\pm$ std\%) of different methods on the LFW face database.}
\label{table_example}
\centering
\begin{tabular}{|c|c|c|c|c|c|c|c|c|c|c|}
\hline
Train No. & LRC& CRC & ProCRC & DLSR & ReLSR & GReLSR & RLRLR & RLSL & LRDLSR(ours)\\
\hline
5 & 29.99$\pm$2.21 & 31.67$\pm$1.16 & 33.19$\pm$0.99 & 30.43$\pm$1.38 & 31.43$\pm$1.13 & 36.76$\pm$1.37 & 36.21$\pm$1.60 & 36.10$\pm$1.82 & \textbf{37.20$\pm$1.66}\\
\hline
6 & 32.37$\pm$1.36 & 34.27$\pm$1.04 & 35.90$\pm$0.93 & 32.35$\pm$1.62 & 34.46$\pm$1.51 & 39.22$\pm$0.92 & 39.37$\pm$1.65 & 38.48$\pm$1.59 & \textbf{39.99$\pm$1.22}\\
\hline
7 & 35.53$\pm$1.69 & 35.96$\pm$1.40 & 36.87$\pm$1.55 & 34.67$\pm$2.45 & 37.50$\pm$2.61 & 43.02$\pm$2.19 & 42.03$\pm$1.42 &41.43$\pm$1.58& \textbf{43.82$\pm$1.23}\\
\hline
8 & 36.98$\pm$1.82 & 37.92$\pm$1.50 & 38.24$\pm$1.15 & 36.27$\pm$1.65 & 38.72$\pm$1.22 & 44.39$\pm$1.77 & 43.30$\pm$1.59 &42.18$\pm$1.37& \textbf{44.88$\pm$1.58}\\
\hline
\end{tabular}
\end{table*}

\subsection{Experiments for Object Classification}
In this section, we validate the performance of our LRDLSR model on the COIL-20 object dataset which has 1440 images of 20 classes. Each class consists of 72 images that are collected at pose intervals of 5 degrees. Some images from this database are shown in Fig. 2. In our experiments, all images are resized to $32\times32$ pixels. For each class, we randomly choose 10, 15, 20, 25 samples to train the model and treat all the remaining images as the test set. The average classification accuracies are reported in Table \uppercase\expandafter{\romannumeral1}. As shown in Table \uppercase\expandafter{\romannumeral1}, we find that our LRDLSR algorithm achieves much better classification results than all the remaining methods used in the comparison, which proves the effectiveness of LRDLSR for the object classification tasks. 

%图2 COIL_20 images
\begin{figure}[!h]
\centering
\includegraphics[scale=1]{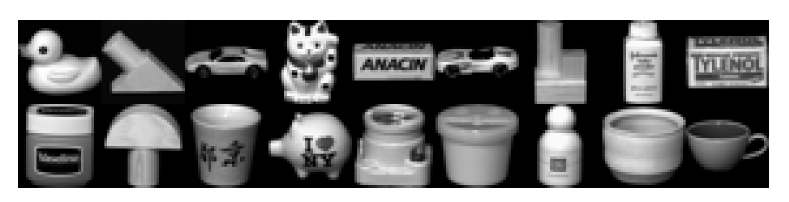}
\caption{Some images from the COIL-20 object database.}
\end{figure}

%图3 face databases images
\begin{figure*}[]
\centering
\includegraphics[scale=0.88]{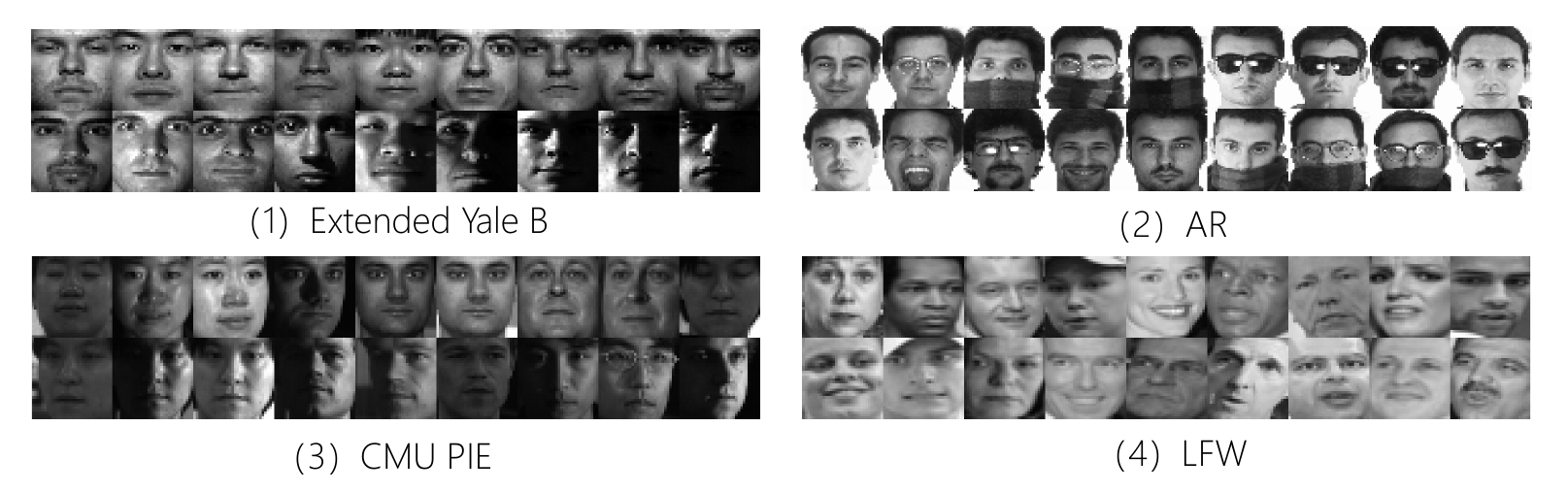}
\caption{Some images from face datasets.}
\end{figure*}

\subsection{Experiments for Face Classification}
In this section, we evaluate the classification performance of LRDLSR on four real face datasets.

(1) The \emph{Extended Yable B Dataset}: The Extended Yale B database consists of 2414 face images of 38 individuals. Each individual has about 59-64 images. All images are resized to 32$\times$32 pixels in advance. We randomly select 10, 15, 20, and 25 images of each individual as training samples, and set the remaining images as test samples.

(2) The \emph{AR Dataset}: We select a subset which consists of 2600 images of 50 women and 50
men and use the projected 540-dimensional features provided in \cite{LCKSVD}. In each individual, we randomly select 3, 4, 5, and 6 images as training samples and the remaining images are set as test samples.

(3) The \emph{CMU PIE Dataset}: We select a subset of this dataset where each individual has 170 images that are collected under five different poses (C05, C07, C09, C27 and C29). All images are resized to 32$\times$32 pixels.  We randomly select 10, 15, 20, and 25 images of each individual as training samples, and treat the remaining images as test samples.

(4) The \emph{LFW Dataset}: Similar to \cite{ICS_DLSR}, we use a subset of this dataset which consists of 1251 images of 86 individuals to conduct experiments. Each individual has 11-20 images. In our experiments, all images are resized to 32$\times$32. We randomly select 5, 6, 7, and 8 images from each individual as training samples and use the remaining images to test. Some images from the above four face databases are shown in Fig. 3.

%图4 tsne_features of LRDLSR

\begin{figure*}[!h]
\centering
\includegraphics[scale=0.45]{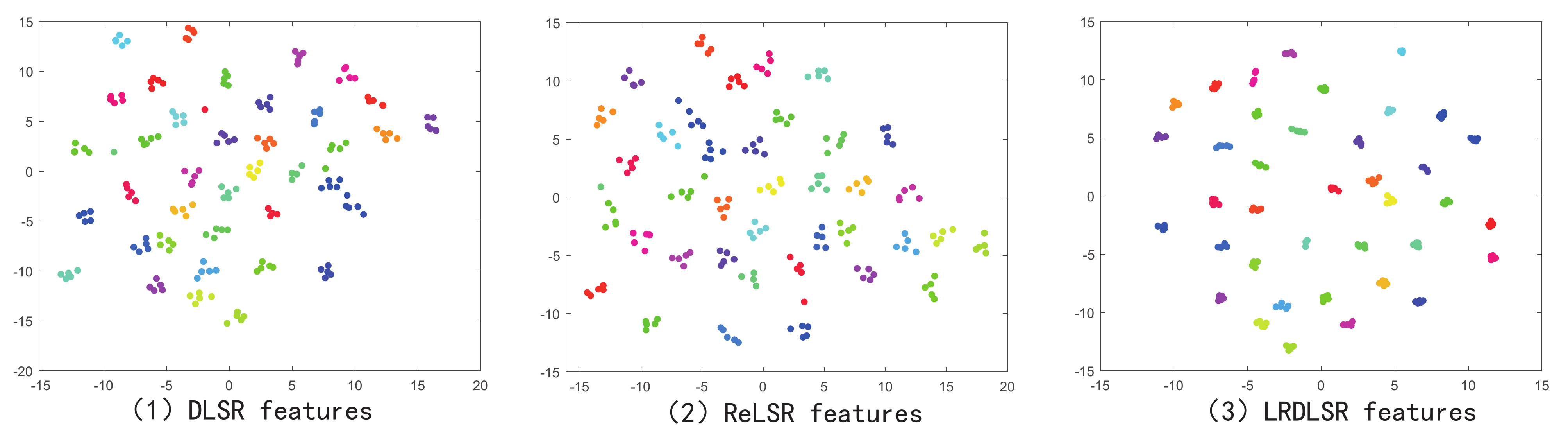}
\caption{T-SNE visualization results of the features extracted by different algorithms on the Extended Yale B database.}
\end{figure*}

%表6 
\begin{table*}[!t]
\renewcommand{\arraystretch}{1.3}
\caption{Recognition accuracies (mean$\pm$ std\%) of LRDLSR and LRDLSR without low-rank constraint.}
\label{table_example}
\centering
\begin{tabular}{|c|c|c|c|c|c|c}
\hline
Database & AR (6) & EYB (15) & CMU PIE (15) & LFW (8) & COIL-20 (15) \\
\hline
LRDLSR & \textbf{92.21$\pm$0.54} & \textbf{94.67$\pm$0.87} & \textbf{94.64$\pm$0.28} & \textbf{45.56$\pm$0.63} & \textbf{97.78$\pm$0.85} \\
\hline
LRDLSR($\beta=0$) & 86.77$\pm$1.25 & 93.39$\pm$0.53 & 90.56$\pm$0.32 & 36.50$\pm$1.60 & 94.48$\pm$0.99 \\
\hline
\end{tabular}
\end{table*}

%图5 parameter analysis of LRDLSR

\begin{figure*}[]
\centering
\includegraphics[scale=0.42]{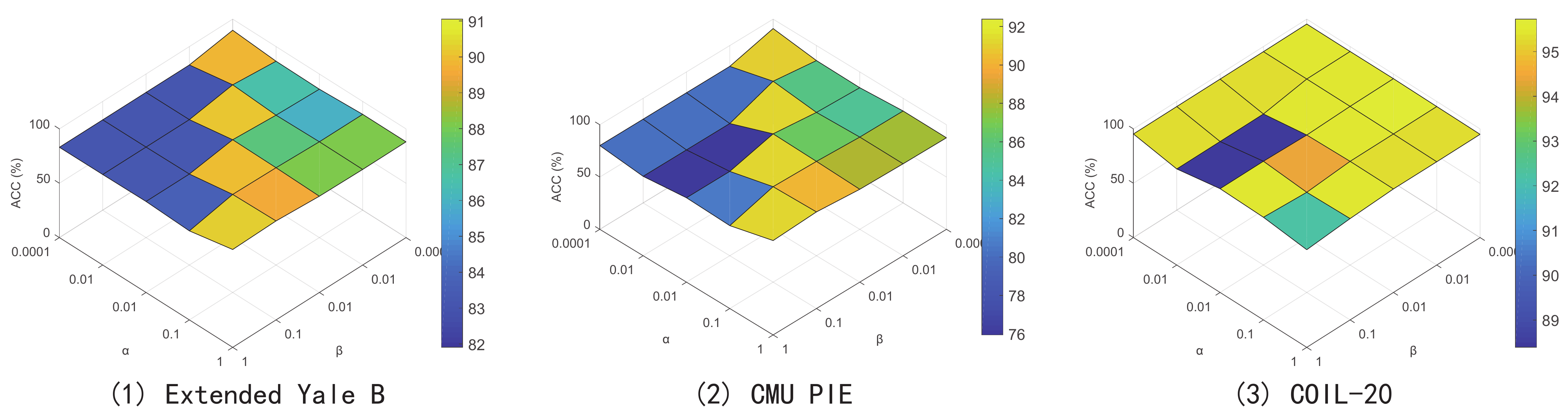}
\caption{The performance evaluation (\%) of LRDLSR versus parameters $\alpha$ and $\beta$ on three different datasets where randomly selected 10 samples for each class are used to train our model.}
\end{figure*}

The average classification rates on these four face datasets are reported in Tables \uppercase\expandafter{\romannumeral2}-\uppercase\expandafter{\romannumeral5}, respectively. It can be observed that our LRDLSR outperforms all the other algorithms on the four face datasets.  The main reason is that our LRDLSR can simultaneously guarantee the intra-class compactness and the inter-class irrelevance of slack regression targets so that more discriminative information is preserved during the projection learning. It is worth noting that the standard deviation of accuracies of LRDLSR are also competetive which demonstrates the robustness of LRDLSR. Besides, we find that the performance gain of LRDLSR is significant when the number of training samples per subject is small, which indicates our model is applicable to small-sample-size problems. Fig. 4 shows the t-SNE \cite{tsne} visualization of the features on the Extended Yale B dataset which are extracted by DLSR, ReLSR and LRDLSR, respectively. We randmly select 5 samples for each individual to validate. It is obvious that the features extracted by LRDLSR model present ideal inter-class seperability and intra-class compactness which is favorable to classification. 

In order to verify whether the low-rank constraint is useful, we set the parameter $\lambda=0$, then test its classification performance. We randomly select 6, 15, 15, 8, and 15 samples per class as the training samples, from the AR, Extended Yale B, CMU PIE, LFW and COIL-20 database, and the remaining samples are treated as test samples. We repeat all experiments ten times and report the average results. The comparative results are shown in Table \uppercase\expandafter{\romannumeral6}.  It is apparent that for $\beta=0$, the classification performance is degraded. Expecially on the LFW database, the difference is actually more than 9$\%$, which indicates that pursuing low-rank intra-class regression targets is indeed helpful to classification.

\subsection{Classification using deep features}
In this section, we conduct experiments on the COIL-20, CMU PIE and LFW databases to further verify whether our model is also effective for deep features. In our experiments, two deep networks, VGG16 \cite{VGG} and ResNet50 \cite{ResNet}, are used. After obtaining the deep features of the original samples, since the dimensionality of features is very high, we first conduct a dimensionality reduction by using PCA so that 98\% of the energy of features is preserved. For the CMU PIE and COIL-20 databases, we randomly select 10 samples of each class for training and all the remaining samples are used for testing. For the LFW database, we randomly select 5 samples of each class for training. Similarly, we repeat all the experiments ten times and report the mean accuracy and standard deviation (mean$\pm$std) of the different algorithms. The experimental results are shown in Table VII. We see that both VGG and ResNet features can achieve better classification accuracy than the original features. Especially on the LFW database, there is nearly a 20$\%$ improvement. Our LRDLSR model with deep features is consistently superior to other algorithms which means that LRDLSR is also appropriate for the deep features.

%表7 
\begin{table}[]
\renewcommand{\arraystretch}{1.3}
\caption{Classification accuracies (mean$\pm$ std\%) on the deep features of the COIL-20, CMU PIE and LFW databases.}
\label{table_example}
\centering
\begin{tabular}{|c|c|c|c|}
\hline
Database & COIL-20 (10) & CMU PIE (10) & LFW (5) \\
\hline
LRDLSR (ours) & 95.12$\pm$1.22 & 91.57$\pm$0.48 & 37.20$\pm$1.66  \\

VGG+LRDLSR (ours) & \textbf{98.65$\pm$1.09} & 91.74$\pm$0.47 & 55.48$\pm$1.55  \\

ResNet+LRDLSR (ours) & 98.59$\pm$0.63 & \textbf{92.98$\pm$0.54} & \textbf{56.19$\pm$1.06}  \\
\hline
RLSL & 94.80$\pm$1.16 & 87.70$\pm$0.63 & 36.10$\pm$1.82  \\

VGG+RLSL & 97.44$\pm$0.72 & 89.05$\pm$0.46 & 53.89$\pm$2.12  \\

ResNet+RLSL & \textbf{97.61$\pm$1.31} & \textbf{89.69$\pm$0.48} & \textbf{54.10$\pm$1.22}  \\
\hline
RLRLR & 92.61$\pm$1.04 & 91.15$\pm$0.58 &  36.21$\pm$1.60 \\

VGG+RLRLR & \textbf{98.60$\pm$0.58} & 91.55$\pm$0.45 &  55.15$\pm$1.47 \\

ResNet+RLRLR & 98.40$\pm$0.67 & \textbf{93.66$\pm$0.40} &  \textbf{55.43$\pm$1.55} \\
\hline
GReLSR & 90.98$\pm$1.62 & 86.88$\pm$0.72 & 36.76$\pm$1.37  \\

VGG+GReLSR & \textbf{97.79$\pm$0.86} & 87.04$\pm$0.63 & 52.18$\pm$1.57  \\

ResNet+GReLSR & 97.73$\pm$0.67 & \textbf{89.87$\pm$0.37} &  \textbf{52.85$\pm$1.53} \\
\hline
ReLSR & 93.65$\pm$1.94 & 88.18$\pm$0.79 &  31.43$\pm$1.13 \\

VGG+ReLSR & 96.90$\pm$1.06 & 88.77$\pm$0.41 & 51.88$\pm$1.42  \\

ResNet+ReLSR & \textbf{96.92$\pm$0.89} & \textbf{89.84$\pm$0.53} &  \textbf{52.91$\pm$1.75} \\
\hline
DLSR & 93.27$\pm$1.43 & 87.54$\pm$0.79 & 30.43$\pm$1.38  \\

VGG+DLSR & \textbf{96.84$\pm$1.43} & 87.47$\pm$0.82 & 49.84$\pm$1.95  \\

ResNet+DLSR & 96.70$\pm$1.65 & \textbf{89.66$\pm$0.63} & \textbf{52.07$\pm$1.91} \\
\hline
\end{tabular}
\end{table}

\subsection{Parameter Sensitivity Validation}
Up to now, it is still an unresolved problem to select optimal parameters for different datasets. In this section, we conduct a sensitivity analysis of the parameters of our LRDLSR model. Note that there are four parameters, i.e., $\alpha$, $\beta$, $\gamma$ and $\lambda$ to be selected in LRDLSR. Among them, $\alpha$ and $\beta$ are respectively used to balance the weight of the slack targets learning term and the class-wise low-rank targets learning term, $\gamma$ and $\lambda$ are respectively used to avoid the overfitting problem of learned targets $T$ and the projection matrix $Q$. For convenience, we set the parameters $\gamma$ and $\lambda$ to 0.01 in advance and focus on selecting the optimal values of parameters $\alpha$ and $\beta$ from the candidate set \{0.0001, 0.001, 0.01, 0.1, 1\} by cross-validation. The classification accuracy as a function of different parameter values on three datasets is shown in Fig. 5. It is apparent that the optimal parameters are different on the respective datasets, but our LRDLSR model is not very sensitive to the values of $\alpha$ and $\beta$. This also demonstrates that compact and similar intra-class targets are critical to discriminative projection learning, but the classification performance does not completely depend on the choice of the parameters.

\section{Conclusion}
In this paper, we proposed a low-rank discriminative least squares regression (LRDLSR) model for multi-class image classification. LRDLSR aims at improving the intra-class similarity of the regression targets learned by the $\varepsilon$-dragging technique. This can ensure that the learned targets are not only relaxed but also discriminative, thus leading to more effective projection. Besides, LRDLSR introduces an extra regularization term to avoid the problem of overfitting by restricting the energy of learned regression targets. The experimental results on the object and face databases demonstrate the effectiveness of the proposed method.
\section*{Acknowledgment}
This work is supported by the National Natural Science
Foundation of China under Grant Nos. 61672265, U1836218,
the 111 Project of Ministry of Education of China under
Grant No. B12018, and UK EPSRC Grant EP/N007743/1,
MURI/EPSRC/DSTL GRANT EP/R018456/1.

\ifCLASSOPTIONcaptionsoff
  \newpage
\fi

\end{document}